\documentclass[letterpaper]{article} % DO NOT CHANGE THIS
\usepackage{aaai25}  % DO NOT CHANGE THIS
\usepackage{times}  % DO NOT CHANGE THIS
\usepackage{helvet}  % DO NOT CHANGE THIS
\usepackage{courier}  % DO NOT CHANGE THIS
\usepackage[hyphens]{url}  % DO NOT CHANGE THIS
\usepackage{graphicx} % DO NOT CHANGE THIS
\usepackage{natbib}  % DO NOT CHANGE THIS AND DO NOT ADD ANY OPTIONS TO IT
\usepackage{caption} % DO NOT CHANGE THIS AND DO NOT ADD ANY OPTIONS TO IT
\usepackage{algorithm}
\usepackage{algorithmic}
\usepackage{booktabs}
\usepackage{multirow}
\usepackage{amsmath}
\usepackage{xcolor}

\usepackage{newfloat}
\usepackage{listings}
\DeclareCaptionStyle{ruled}{labelfont=normalfont,labelsep=colon,strut=off} % DO NOT CHANGE THIS
\floatstyle{ruled}
\newfloat{listing}{tb}{lst}{}
\floatname{listing}{Listing}
\title{Political Actor Agent: Simulating Legislative System for Roll Call Votes Prediction with Large Language Models}
\author {
    Hao Li,
    Ruoyuan Gong,
    Hao Jiang\thanks{Corresponding author}
}
\affiliations {
    Wuhan University, Wuhan, 430072 China\\
    whulh@whu.edu.cn, GongRuoyuan@whu.edu.cn, jh@whu.edu.cn
}

\begin{document}

\maketitle

\begin{abstract}
  Predicting roll call votes through modeling political actors has emerged as a focus in quantitative political science and computer science. Widely used embedding-based methods generate vectors for legislators from diverse data sets to predict legislative behaviors. However, these methods often contend with challenges such as the need for manually predefined features, reliance on extensive training data, and a lack of interpretability. Achieving more interpretable predictions under flexible conditions remains an unresolved issue. This paper introduces the Political Actor Agent (PAA), a novel agent-based framework that utilizes Large Language Models to overcome these limitations. By employing role-playing architectures and simulating legislative system, PAA provides a scalable and interpretable paradigm for predicting roll call votes. Our approach not only enhances the accuracy of predictions but also offers multi-view, human-understandable decision reasoning, providing new insights into political actor behaviors. We conducted comprehensive experiments using voting records from the 117-118th U.S. House of Representatives, validating the superior performance and interpretability of PAA. This study not only demonstrates PAA's effectiveness but also its potential in political science research.
\end{abstract}


\section{Introduction}
Legislative actions, such as proposing, reviewing, and voting on bills, enable political actors to influence national and societal development. Modeling these actors has emerged as an interdisciplinary focus within quantitative political science and computer science. The representations obtained from modeling political actors are applied to downstream tasks such as roll-call vote prediction \cite{par, align}, and political stance prediction \cite{kgap, news}.

In this paper, we primarily focus on the problem of predicting legislators' roll call votes.With voting records, bill texts, and background knowledge, there are two main approaches to modeling political actors. The ideal point model, one of the most widely used methods for roll call vote prediction, represents legislators and bills as points in one or multiple dimensions \cite{ideal, embedding}. Recently, some studies have employed heterogeneous information graphs to represent legislators, bills, and contextual knowledge, including complex relationships between party affiliations, lobbying \cite{understanding}, and assets \cite{par, kgap, unifying}. These studies then use heterogeneous graph neural networks to generate embeddings for nodes within the graph and predict voting outcomes. Both approaches embed legislators and bills into a vector space and use neural networks or similarity measures to predict results.

However, the aforementioned embedding-based methods exhibit several limitations: 
1. \textbf{Limitations of predefined features: } The model's training relies solely on predefined features, preventing natural extension to new, untrained relationships.
2. \textbf{Volume of training data:} Most models depend on large datasets to achieve optimal performance, which is not feasible in real-world scenarios, such as predicting votes of newly elected legislators.
3. \textbf{Interpretability of predictions:} Predictions based on embeddings lack interpretability, particularly in providing insights in a manner understandable to humans.

\begin{figure}[!ht]
  \centering
  \includegraphics[width=0.9\columnwidth]{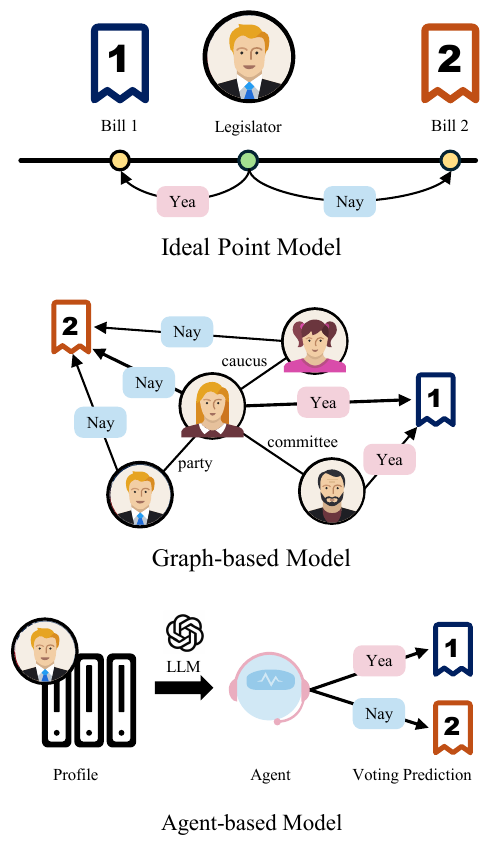} % Reduce the figure size so that it is slightly narrower than the column. Don't use precise values for figure width.This setup will avoid overfull boxes.
  \caption{Examples of different political actor modeling methods include: The ideal point model represents legislators and bill entities as vectors. The graph-based model embeds nodes from heterogeneous information graphs into vectors using a graph embedding model. Our agent-based model does not rely on distances between embeddings; instead, it directly generates voting outcomes using LLM agents.}
  \label{fig1}
\end{figure}

To address these challenges, we have turned our attention to the accomplishments of agent research based on Large Language Models (LLMs) \cite{survey}. With designed profile, planning, and action modules, LLM agents can exhibit intelligent decision-making behaviors. Some studies have applied LLM Agents in fields such as economics \cite{economicus, investors}, social simulation \cite{unveiling, demographics, artificial}, and voting decision-making \cite{voting, aivoting}. As shown in figure \ref{fig1}, by reframing the problem of modeling political actors as constructing political agents, we introduce deeper insights into this field.

Specifically, we designed the Political Actor Agent (PAA) based on a role-playing architecture \cite{camel}, which offers several advantages:
1. \textbf{Scalable Politician Profile:} Each agent is equipped with a scalable profile. This profile is more flexible and easier to manage compared to manually designed relational rules.
2. \textbf{Multi-view Planning:} From various views, such as the delegate and trustee views \cite{representation}, the PAA can formulate different voting plans. Decision-making reasons understandable to humans can better provide new insights for political science research.
3. \textbf{Simulated Legislative Action:} Based on legislators' voting strategies, we developed an Influence Mechanism that simulates parliamentary dynamics. Legislators are categorized as leader agents and follower agents, with follower agents being influenced by the voting outcomes of leader agents. This mechanism allows for accurate vote predictions even with limited known data. Our contributions are as follows:

\begin{enumerate}
  \item We propose a new agent-based paradigm for political actor research. Compared to embedding-based methods, our role-playing framework for legislator simulation offers more accurate and interpretable results for corresponding downstream tasks.
  \item We introduce the Political Actor Agent (PAA) for roll-call vote prediction. Our approach, through the design of scalable profiles, multi-view planning, and simulated legislator actions, significantly enhances prediction accuracy and offers interpretable decision-making insights for political science research.
  \item We conduct comprehensive experiments using voting records from the 117-118th House of Representatives. Our experiments demonstrate that our method not only achieves high prediction accuracy but also provides interpretable political insights.
  \end{enumerate}

\section{Related Works}
This section introduces roll call voting prediction based on political actor modeling and the use of LLM agents for decision simulation.

\subsection{Modeling Political Actors for Voting Prediction}

Legislators' voting behavior in parliament has been a primary research focus due to its characteristic transparency and significance. One of the most popular techniques in political science is the ideal point model, constructed based on voting records and typically used to represent unidimensional or multidimensional ideological positions \cite{spatial, ideal}. The ideal point model has been expanded in several studies. For instance, \cite{gerrish} employ a topic model to perform a detailed analysis of legislative texts, enhancing the contextual understanding of votes. Further extending this approach, researchers have developed a topic factorized ideal point model that assigns ideal points for each topic rather than globally, allowing for more nuanced interpretations of legislative behavior \cite{factorized}. Additionally, efforts have been made to learn multidimensional embeddings of legislators and bills to improve prediction accuracy \cite{embedding}, and to integrate broader political texts, such as speeches and tweets, into the ideal point model to enrich the dataset \cite{textbased}.

In terms of incorporating richer contextual information, graph-based methods, driven by advancements in knowledge graphs and graph neural networks, have gained popularity. External knowledge is introduced into voting prediction in the form of heterogeneous information graphs. Compared to ideal point models, graph-based models can more flexibly capture complex political relationships, such as cosponsorship \cite{joint}, donors \cite{understanding}, and stakeholders \cite{modeling}. Additionally, more expert knowledge has been incorporated, such as news data \cite{kgap}, Twitter statements \cite{align}, wiki pages, and political think tanks \cite{par}.

Political decision-making often involves highly complex mechanisms and background knowledge. Recent studies have focused on leveraging large language models (LLMs). For instance, \cite{unifying} constructed a multi-view political knowledge graph to enhance the domain knowledge of LLMs. Similarly, \cite{uppam} developed a pretraining architecture that maps language to actor representations. These attempts demonstrate the capability of LLMs in related tasks but remain complementary to embedding-based methods. This paper proposes a novel approach to modeling legislators as actors from an agent view, aiming to advance the prediction and understanding of voting behavior.

\subsection{LLM Agent in Decision Simulation}
Roll call voting can be viewed as a decision-making behavior. As LLM agents demonstrate the potential for human-level intelligence \cite{survey}, many studies have applied them in fields such as natural sciences \cite{emergent}, software engineering \cite{chatdev}, and embodied intelligence \cite{embodied}. Some researchers are exploring the integration of large language models with social sciences, particularly in simulating decision-making behaviors. In economics, for instance, \cite{silicus} treat LLM agents as economic agents, observing their economic decisions under different conditions and scenarios. \cite{rational} propose a financial bias indicator framework, analyzing irrational biases in LLM agents through behavioral finance theories.

In the realm of voting decision simulation, studies like \cite{out} investigate the feasibility of using large language models to simulate human samples, designing agents based on demographic data to model the reactions of different populations. The article \cite{voting} finds that the input method and presentation of choices can affect the voting behavior of LLMs. \cite{generative} delve deeper into the issue of bias in the voting process, discovering that equal-share methods can lead to fairer voting outcomes.

The above research on decision simulation indicates that LLM agents can offer new insights into social sciences. Unlike agents based on demographic data, this paper explores a more sophisticated decision simulation. By setting detailed political profiles and voting mechanisms, we aim to predict the roll-call voting results of political actors. Our goal is to investigate the ability of LLM agents to simulate complex, diverse outcomes and to summarize their behavior based on specific political perspectives.

\section{Method}
\begin{figure*}[!ht]
  \centering
  \includegraphics[width=0.76\paperwidth]{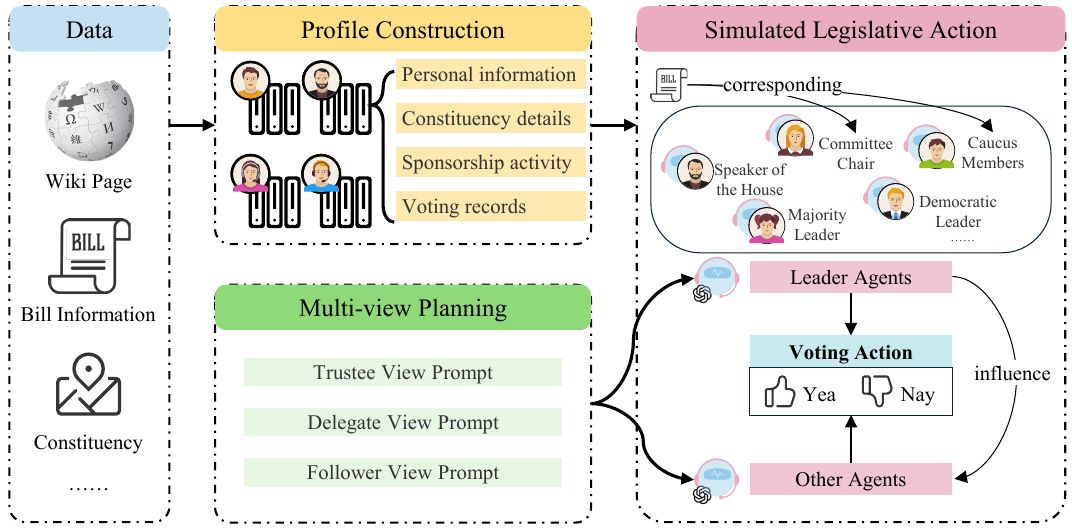} % Reduce the figure size so that it is slightly narrower than the column. Don't use precise values for figure width.This setup will avoid overfull boxes.
  \caption{Framework of PAA}
  \label{fig2}
\end{figure*}

In this section, we detail the framework of the Political Actor Agent (PAA) for roll call vote prediction. As illustrated in figure \ref{fig2}, we begin by collecting data from Wikipedia and legislative bills from the Congress. We then construct scalable agent profiles for each legislator, incorporating personal information, constituency details, and data on bills they have sponsored and voted on. Subsequently, we design a multi-view planning module for each agent to perform reasoning based on various political science views. Finally, in the simulated legislative action module, we implement an influence mechanism that allows leader agents associated with the bill to act first, with the remaining agents making their choices after learning the leaders' decisions. This approach leverages political science perspectives and LLM agents to achieve more accurate and interpretable voting predictions.

\subsection{Profile Construction Module}
Using a role-playing architecture, we simulate characters where agents assume specific roles to make voting decisions. The profile construction module is integrated into the prompt to influence the design and behavior of the LLMs. The design of the profile is highly scalable and can synthesize information from different data sources under various conditions, such as personal basic information, career history, voting records, etc. Specifically, the profiles we use include the following personal details:

\subsubsection{Personal information} This refers to the basic information of legislators such as party affiliation, committee memberships, core group affiliations, educational background, number of children, place of birth, and other relevant details.

\subsubsection{Constituency details} This section details the legislator's constituency, including information such as the median family income of the district, urban-rural population distribution, and total population. This information helps model the social and economic context in which the legislator operates.

\subsubsection{Sponsorship activity} Each bill is initiated by a sponsor and possibly several cosponsors. Historical records of bill sponsorship are extremely useful for modeling a legislator’s specific preferences. Past studies have also demonstrated the significance of sponsorship information in political actor modeling \cite{joint}.

\subsubsection{Voting records:} Historical voting records provide the most direct data on a legislator's political stance. Unlike previous political actor models, the PAA integrates past voting records directly into the profile without explicit model training. Leveraging the capabilities of large language models, we can predict voting outcomes based on a limited amount of known information, which is particularly useful when modeling newly elected legislators with less data available.

\subsection{Multi-view Planning Module}
Inspired by political science, the planning module decomposes the task of voting decision-making into three main views, aiding the agent in decision-making. In the appendix, we present the prompts used for these different views \cite{representation}.

\subsubsection{Trustee view}  The Trustee view implies that the legislator relies on their expertise, making decisions based on what they believe to be the best policies for their constituents and the country.

\subsubsection{Delegate view} The Delegate view indicates that the legislator sees their role as expressing the will of the majority of their constituents, acting in accordance with the wishes of their voters.

\subsubsection{Follower view} Legislators in this category follow the opinions of their party leaders. They are not keen on reflecting public opinion directly and often lack substantial personal insights.

Practical evidence shows that legislators' decisions are influenced by multiple views. After planning through these various views, the agent synthesizes these results to arrive at the final decision.

\subsection{Simulated Legislative Action Module}
Previous political actor models, while recognizing differences among legislators, typically generated predictions in a single step. Embedding-based methods fail to intricately simulate the decision-making processes of real legislators and do not effectively model the influence of leader figures. In the PAA, we have designed an "influence mechanism" to model how leading agents influence other agents.

\subsubsection{Leader agents} Leader agents, denoted as \(L\) , comprise the following agents:
\begin{equation}
  L = \{ S, R, D, CC, CM\},
\end{equation}
where \(S\) represents the Speaker of the House, \(R\) and \(D\) represent the Republican and Democratic Leader, respectively, with information sourced from the official website \footnote{\url{https://www.house.gov/leadership}}. \(CC\) denotes the Chairperson of the committee introducing the bill, and \(CM\) represents caucus members related to the bill. A congressional caucus is a group of members of the United States Congress who meet to pursue common legislative objectives, often actively pushing legislation through their actions.

\subsubsection{Influence mechanism} In the influence mechanism, leader agents first vote based on the multi-view planning module.
\begin{equation}
  V_{l} = p(L),
\end{equation}
where \(V_{l}\) represents the voting prediction of leader agents, and \(p\) represents the multi-view planning module. The voting prediction for the remaining agents \(O\) is as follows:
\begin{equation}
  V_{o} = p(O|V_{l}),
\end{equation}
where \(V_{o}\) represents the voting predictions of other agents, made under the condition of knowing \(V_{l}\), \(p(O|V_{l})\) indicates that we include 
\(V_{l}\) in the prompt to predict the vote of agent \(O\). This method effectively approximates the real-world legislative process. It is worth noting that this influence mechanism is highly flexible and can be adapted to different conditions for selecting leadership agents. The configuration presented here is just one of the possible scenarios.

\section{Experiments}
In this section, we conduct a detailed evaluation of the Political Actor Agent (PAA) in predicting roll call votes. Section 4.1 introduces the datasets selected for the experiment and the baselines used. Section 4.2 details the experimental results across different dataset splits. Sections 4.3 and 4.4 feature extensive ablation studies, analyzing the impact of various PAA modules on the outcomes, with a deep dive into the profile module. Section 4.5 demonstrates the consistency of the PAA's results. Section 4.6 provides an interpretable example to illustrate how the PAA generates its predictions.

\subsection{Datasets and Baselines}
We selected voting data from the 117th to 118th House of Representatives, covering 432 legislators. In addition to the voting data for bills, we collected additional data for constructing profiles and heterogeneous information graphs. This includes the most recent Wikipedia pages of the legislators (as of March 2024), Wikipedia pages for all constituencies, data on the sponsors and cosponsors of bills, and Twitter posts by the legislators.
We selected five methods as baselines, including a variant of the ideal point model, three graph-based models, and one that incorporates a pre-trained model.
\begin{enumerate}
\item \textbf{ideal-vector} \cite{embedding}: A multi-dimensional ideal point model based on word embeddings, learning politician representations from bill texts.
\item \textbf{LSTM+GCN} \cite{joint}: A graph-based model using a Graph Convolutional Network (GCN) to generate representations of legislators and an LSTM to generate representations of bill texts.
\item \textbf{Vote+MTL} \cite{align}: A graph-based model that incorporates Twitter data, using a Relational Graph Convolutional Network (RGCN) to generate representations of politicians.
\item \textbf{PAR} \cite{par}: A graph-based model that combines various socio-contextual information, integrating representational models with expert knowledge to generate politician representations.
\item \textbf{UPPAM} \cite{uppam}: A contrastive learning framework based on the social network of legislators and bill texts, using pre-trained models to generate politician representations.
\end{enumerate}

\subsection{Roll Call Vote Prediction}
\subsubsection{Implementation}
We divided the dataset in chronological order and selected three different ratios for dataset splits. 
\begin{enumerate}
\item \(\text{split}_{244}\): 20\% training, 40\% validation, 40\% testing. 
\item \(\text{split}_{433}\): 40\% training, 30\% validation, 30\% testing. 
\item \(\text{split}_{622}\): 60\% training, 20\% validation, 20\% testing.
\end{enumerate}

As an agent-based approach, the PAA does not train models. Instead, we sample 20 voting records from the training set to construct the profile for each agent and directly evaluate the results on the test set.

Our methods were tested using Llama-3-70B (\(\text{PAA}_{\text{L}}\)) and GPT-4o-mini (\(\text{PAA}_{\text{G}}\)) as the base models. We chose accuracy and macro-averaged F1-score as metrics to evaluate the model's performance on a three-class task: in favor, against, and abstaining. We conducted experiments for \(\text{PAA}_{\text{L}}\) on a four NVIDIA RTX A6000 GPUs, while the \(\text{PAA}_{\text{G}}\) experiments were carried out using the OpenAI API\footnote{\url{https://platform.openai.com/docs/api-reference/introduction}}.

\begin{table*}[ht]
  \centering
  \begin{tabular}{lcccccc}
    \hline
    \multirow{2}{*}{Method} & \multicolumn{2}{c}{\(\text{split}_{244}\)} & \multicolumn{2}{c}{\(\text{split}_{433}\)} & \multicolumn{2}{c}{\(\text{split}_{622}\)} \\ 
    \cmidrule(lr){2-3} \cmidrule(lr){4-5} \cmidrule(l){6-7}
    & acc & f1 & acc & f1 & acc & f1 \\
    \hline
    ideal-vector & 80.9 \(\pm\) 0.12 & 79.2 \(\pm\) 0.13 & 82.2 \(\pm\) 0.23 & 80.6 \(\pm\) 0.36 & 86.1 \(\pm\) 0.32 & 84.4 \(\pm\) 0.25 \\
    LSTM+GCN & 83.5 \(\pm\) 0.13 & 82.5 \(\pm\) 0.12 & 85.3 \(\pm\) 0.12 & 83.2 \(\pm\) 0.35 & 87.8 \(\pm\) 0.08 & 85.2 \(\pm\) 0.12 \\
    Vote+MTL & 83.2 \(\pm\) 0.04 & 83.5 \(\pm\) 0.12 & 84.2 \(\pm\) 0.20 & 84.3 \(\pm\) 0.03 & 89.5 \(\pm\) 0.14 & 86.2 \(\pm\) 0.06 \\
    PAR & 85.3 \(\pm\) 0.11 & 80.2 \(\pm\) 0.12 & 85.8 \(\pm\) 0.12 & 85.2 \(\pm\) 0.12 & 90.2 \(\pm\) 0.23 & 87.5 \(\pm\) 0.03 \\
    UPPAM & \underline{86.5 \(\pm\) 0.10} & 80.5 \(\pm\) 0.07 & \underline{88.5 \(\pm\) 0.03} & 85.9 \(\pm\) 0.03 & \underline{91.7 \(\pm\) 0.08} & 86.3 \(\pm\) 0.09 \\
    \(\text{PAA}_{\text{L}}\) & 85.9 \(\pm\) 0.10 & \underline{87.2 \(\pm\) 0.08} & 85.7 \(\pm\) 0.10 & \underline{88.1 \(\pm\) 0.07} & 87.7 \(\pm\) 0.05 & \underline{89.6 \(\pm\) 0.05} \\
    \(\text{PAA}_{\text{G}}\) & \textbf{91.8 \(\pm\) 0.15} & \textbf{92.2 \(\pm\) 0.10} & \textbf{91.3 \(\pm\) 0.20} & \textbf{91.7 \(\pm\) 0.12} & \textbf{92.1 \(\pm\) 0.10} & \textbf{93.0 \(\pm\) 0.12} \\
    \hline
  \end{tabular}
  \caption{Results of various methods on different splits. The best results are highlighted in bold, and the second-best results are underlined. Each experimental group was run five times, and we show the mean and standard deviation in the table.}
  \label{tab1}
\end{table*}

\subsubsection{Results}
The results are presented in table \ref{tab1}. The primary findings of the experiment indicate that \(\text{PAA}_{\text{G}}\) consistently outperformed across all dataset splits, achieving superior results compared to state-of-the-art methods. \(\text{PAA}_{\text{L}}\) performed particularly well in terms of the macro-averaged F1 score. This is largely due to the label imbalance in the three-class voting problem. As the proportion of the training set decreases within the dataset, the performance of embedding-based methods declines more rapidly, whereas PAA remains more stable. This suggests that PAA is better adapted to scenarios with limited data, such as predicting the voting behavior of new legislators.

PAR and UPPAM performed next best to \(\text{PAA}_{\text{G}}\) in the \(\text{split}_{622}\). PAR learns representations of legislators through external knowledge and social media corpora, while UPPAM utilizes extensive social media data to train a pre-trained model. Their performance in the voting prediction task also validates the capabilities of large-scale external knowledge and pre-trained models. Unlike these methods, our approach does not utilize any social media data. With the aid of LLMs, PAA achieves comparable or even superior results under less time and data constraints.

\subsection{Ablation Studies}
\subsubsection{Implementation}
We designed ablation experiments to further verify the impact of various modules on the performance of the PAA. On the \(\text{split}_{244}\), while keeping other conditions constant, we individually removed the profile module, the planning module, and the acting module to assess their individual contributions to the PAA's effectiveness.
\begin{enumerate}
  \item PAA w/o Pro: Remove profile module.
  \item PAA w/o Pla: Remove planning module.
  \item PAA w/o Act: Remove acting module.
\end{enumerate}

\subsubsection{Result}
The experimental results are shown in table \ref{tab2}. The results indicate that different modules impact the performance of the PAA to varying degrees. Notably, the profile module has the greatest effect, as it contains a wealth of crucial information, including personal details, voting records, and more. Next, we will further explore the influence of the profile on the performance of the PAA.
\begin{table}[ht]
  \centering
  \begin{tabular}{lcc}
    \hline
    \multirow{2}{*}{Method} & \multicolumn{2}{c}{\(\text{split}_{244}\)}  \\ 
    \cmidrule(lr){2-3} 
    & acc & f1 \\
    \hline
    PAA w/o Pro & 78.6 \(\pm\) 0.10 & 73.4 \(\pm\) 0.07   \\
    PAA w/o Pla & 80.1 \(\pm\) 0.12 & 74.1 \(\pm\) 0.13  \\
    PAA w/o Act & 80.7 \(\pm\) 0.19 & 74.2 \(\pm\) 0.16   \\
    \(\text{PAA}_{\text{G}}\) & 91.8 \(\pm\) 0.15 & 92.2 \(\pm\) 0.10   \\
    \hline
  \end{tabular}
  \caption{The results of ablation experiments conducted on the profile, planning, and action modules.}
  \label{tab2}
\end{table}

\subsection{Analysis of Profile Module}
\subsubsection{Implementation}
The profile module comprises four key components: legislator personal information, constituency details, legislative sponsorship activity, and voting records. We have designed experiments to address specific concerns related to the performance of the PAA. \textbf{RQ1}: The outstanding performance of PAA might be attributed to the fact that the agent, based on a large language model, had already been exposed to legislative information and legislator-related corpora during the pre-training phase. \textbf{RQ2}: Which part of the agent profile plays a more significant role in predicting outcomes? \textbf{RQ3}: Does an excessive length of voting records dilute the importance of other components and adversely affect the performance of the experiments?

Based on the considerations above, for RQ1 and RQ2, we conducted detailed experiments on the \(\text{split}_{244}\) dataset with \(\text{PAA}_{\text{G}}\). For RQ3, we devised an experiment to analyze the effect of the length of voting records on the performance of the PAA. One approach involves sampling 20 data points from the training set to add to the profile, while another uses the entire training set. The PAA variants used in the experiment are as follows:
\begin{enumerate}
  \item \(\text{PAA}_{\text{ano}}\): Anonymize the names of legislators and bill numbers. Names and bill numbers are replaced with random numbers.
  \item \(\text{PAA}_{\text{dec}}\): Introduce incorrect information to deceive the large language model; we swapped names of legislators with differing stances to observe whether PAA predicts based on pre-trained data.
  \item PAA w/o Per: Remove basic information about legislators from the profile.
  \item PAA w/o Con: Remove constituency information from the profile.
  \item PAA w/o Spo: Remove information about Sponsored and Cosponsored Bills from the profile.
  \item PAA w/o Rec: Remove voting records from the profile.
  \item \(\text{PAA}_{\text{G}}^{\text{F}}\), \(\text{PAA}_{\text{L}}^{\text{F}}\): Use the entire training set data for voting in the profile.
\end{enumerate}

\subsubsection{Results}
The results are presented in table \ref{tab3}. From the findings, we can see that both \(\text{PAA}_{\text{ano}}\) and \(\text{PAA}_{\text{dec}}\) exhibit a decrease in predictive performance when names and bill identifiers are modified. The fact that \(\text{PAA}_{\text{dec}}\) performs slightly worse than \(\text{PAA}_{\text{ano}}\) suggests that legislator information has a certain impact on the experimental results. Yet, both variants still significantly outperform the current baselines. Our findings indicate that model performance is only slightly affected by the names, suggesting that PAA likely relies on the information in our profile module for predictions.

Additionally, the study of the profile reveals that all four types of information we selected are effective for prediction. Basic information, sponsorship, and voting records have the most significant impact, while constituency information has the least. This might be because the relationship between constituency and voting behavior is less apparent compared to other information, suggesting the need for more explicit mechanisms to better utilize this data.

\begin{table}[ht]
  \centering
  \begin{tabular}{lcc}
    \hline
    \multirow{2}{*}{Method} & \multicolumn{2}{c}{\(\text{split}_{244}\)}  \\ 
    \cmidrule(lr){2-3} 
    & acc & f1 \\
    \hline
    PAA-ano & 90.8 \(\pm\) 0.10 & 91.3 \(\pm\) 0.08   \\
    PAA-Dec & 90.1 \(\pm\) 0.08 & 90.7 \(\pm\) 0.11  \\
    PAA w/o Per & 82.9 \(\pm\) 0.17 & 76.3 \(\pm\) 0.23   \\
    PAA w/o Con & 90.9 \(\pm\) 0.06 & 91.3 \(\pm\) 0.09   \\
    PAA w/o Spo & 83.6 \(\pm\) 0.09 & 78.9 \(\pm\) 0.05   \\
    PAA w/o Rec & 83.4 \(\pm\) 0.13 & 77.8 \(\pm\) 0.19   \\
    \(\text{PAA}_{\text{G}}\) & 91.8 \(\pm\) 0.15 & 92.2 \(\pm\) 0.10   \\
    \hline
  \end{tabular}
  \caption{The results of the ablation experiments that modified the profile module.}
  \label{tab3}
\end{table}

\begin{figure}[!ht]
  \centering
  \includegraphics[width=0.9\columnwidth]{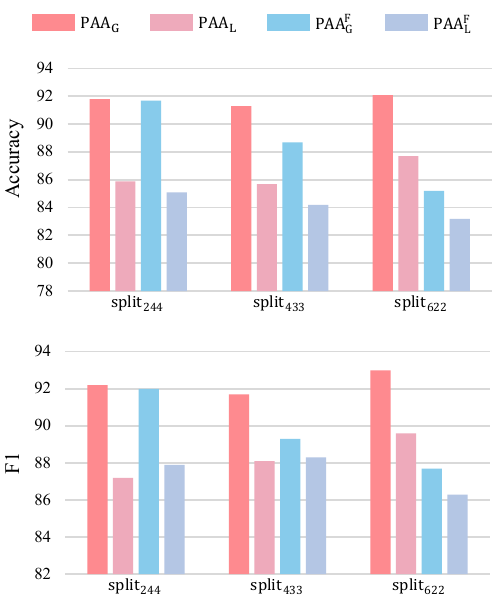} % Reduce the figure size so that it is slightly narrower than the column. Don't use precise values for figure width.This setup will avoid overfull boxes.
  \caption{The results on the impact of profile length on the performance of the PAA.}
  \label{fig3}
\end{figure}

The experimental results, as shown in figure \ref{fig3}, indicate that as the size of the training set increases, the performance of \(\text{PAA}_{\text{L}}\) and \(\text{PAA}_{\text{G}}\) improves, achieving the best results on the \(\text{split}_{622}\) dataset. However, the performance of \(\text{PAA}_{\text{L}}^{\text{F}}\) and \(\text{PAA}_{\text{G}}^{\text{F}}\) gradually declines, with the worst results observed on the \(\text{split}_{622}\) dataset. This suggests that an excessive number of voting records can confuse the large language model, preventing it from capturing other important information. Therefore the PAA opts for a sampling method to construct profiles, achieving better performance with fewer resources.

\subsection{Consistency Analysis}
\subsubsection{Implementation}
Unlike other embedding-based models that generate representations of legislators, large language models are prone to underlying hallucinations, potentially generating different results across multiple runs. Therefore, we analyzed the consistency of results produced by the PAA. We randomly selected 50 legislator agent and bill pairs and repeated the experiment 20 times.

\subsubsection{Results}
\begin{figure}[!ht]
  \centering
  \includegraphics[width=0.9\columnwidth]{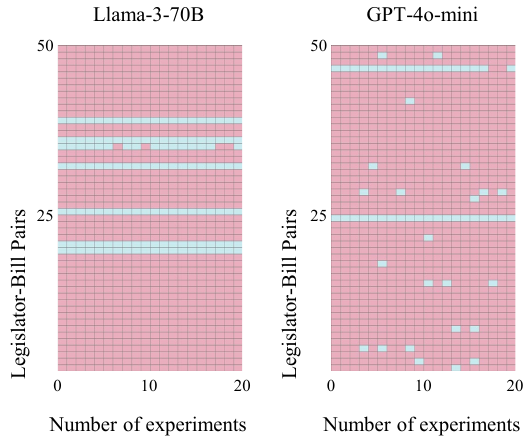} % Reduce the figure size so that it is slightly narrower than the column. Don't use precise values for figure width.This setup will avoid overfull boxes.
  \caption{The consistency experiment results.}
  \label{fig4}
\end{figure}

The experimental results using \(\text{PAA}_{\text{G}}\) and \(\text{PAA}_{\text{L}}\) are shown in figure \ref{fig4} respectively. In the heatmaps, red indicates correct predictions, while blue denotes incorrect predictions. The horizontal axis represents different experimental runs, and the vertical axis represents different agent-bill pairs. Overall, we observed that the hallucination phenomenon commonly associated with large language models was not significant. Specifically, \(\text{PAA}_{\text{L}}\) showed stronger consistency, with only a few instances where the same agent-bill pair produced different results across experiments. Meanwhile, \(\text{PAA}_{\text{G}}\) achieved a higher number of correct predictions, though its consistency was not as strong as that of Llama-3-70B. Thus, conducting multiple experiments to mitigate the effects of hallucinations is highly beneficial.

\subsection{Interpretability}
\begin{figure}[!ht]
  \centering
  \includegraphics[width=0.9\columnwidth]{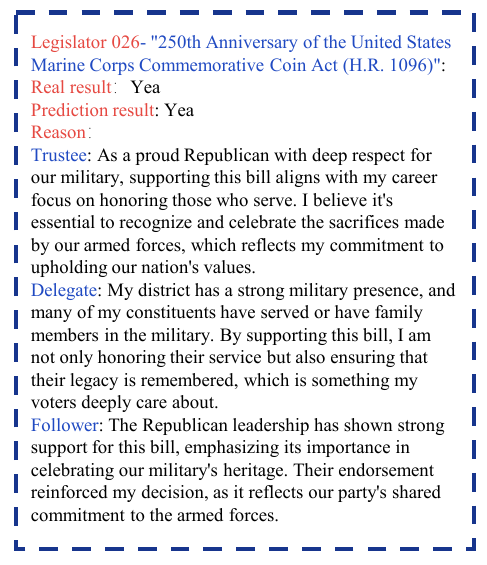} % Reduce the figure size so that it is slightly narrower than the column. Don't use precise values for figure width.This setup will avoid overfull boxes.
  \caption{An example demonstrating how an agent cast a vote and subsequently summarize the reasons for its decision.}
  \label{fig5}
\end{figure}

As shown in figure \ref{fig5}, we present an example to illustrate the interpretability of the PAA's voting prediction results. In this example, the Agent explains its choice from three different views, each backed by factual evidence. It is noteworthy that he mentioned his conservative inclinations and the corresponding choice of leadership agent, which highlights the effectiveness of our multi-view planning module and influence mechanism.

\section{Conclusion}
In this study, we introduced the Political Actor Agent (PAA), an innovative approach leveraging Large Language Models (LLMs) for the predictive modeling of legislative behavior. By incorporating agents in a role-playing architecture, PAA uniquely simulates the dynamics of legislative decision-making, providing a robust framework for understanding and predicting roll call votes. The utilization of extensive pre-existing knowledge and reasoning capabilities from LLMs ensures high accuracy and interpretability without relying on massive bespoke training datasets. Additionally, although our experiments focused on U.S. legislators, PAA can be easily extended to other countries. 

However, the PAA also has several limitations: 1. \textbf{Data Diversity: } Despite achieving superior performance compared to baselines with fewer data types, the current architecture lacks support for integrating diverse data sources like social media commentary and news. 2. \textbf{Task Diversity: } Compared to existing political actor modeling methods, PAA primarily supports the task of roll call vote prediction. Developing mechanisms to support more downstream tasks remains an unresolved issue. 3. \textbf{Hallucination: } Stable and consistent prediction results are essential, although we have analyzed PAA's results from a consistency perspective, The hallucination issue in LLMs is complex, and consistency is only one measure of it.

In the future, we plan to further explore the integration of more diverse data types, such as real-time social media analytics and global news events, to enhance the predictive power of PAA. Additionally, we aim to design more versatile mechanisms to accommodate different downstream tasks, broadening the applicability and utility of the PAA in computational political science.

\section{Acknowledgments}
This work was supported by the Key R\&D Program of China under Grant 2022YFB2901202.  
\bibliography{aaai25}

\end{document}